\pdfoutput=1
\documentclass[11pt]{article}

\usepackage[final]{acl}
\usepackage{times}
\usepackage{latexsym}

\usepackage[T1]{fontenc}
\usepackage[utf8]{inputenc}

\usepackage{microtype}

\usepackage{inconsolata}

\usepackage{graphicx}

\usepackage{multirow}
\usepackage{mdframed}

\title{Generating Diverse Personas for User Simulators to Test Interview Dialogue Systems}
\author{
  \textbf{Mikio Nakano\textsuperscript{1,2}},
  \textbf{Kazunori Komatani\textsuperscript{2}},
  \textbf{Hironori Takeuchi\textsuperscript{3}}
\\
\\
  \textsuperscript{1}C4A Research Institute, Inc., Setagaya, Tokyo, Japan \\
  \textsuperscript{2}SANKEN, University of Osaka, Ibaraki, Osaka, Japan\\
  \textsuperscript{3}Musashi University,  Nerima, Tokyo, Japan\\
 \texttt{mikio.nakano@c4a.jp, komatani@sanken.osaka-u.ac.jp}\\
 \texttt{h.takeuchi@cc.musashi.ac.jp}
  }

\begin{document}
\maketitle
\begin{abstract}

This paper addresses the issue of the significant labor required to test interview dialogue systems. While interview dialogue systems are expected to be useful in various scenarios, like other dialogue systems, testing them with human users requires significant effort and cost. Therefore, testing with user simulators can be beneficial. Since most conventional user simulators have been primarily designed for training task-oriented dialogue systems, little attention has been paid to the personas of the simulated users. During development, testing interview dialogue systems requires simulating a wide range of user behaviors, but manually creating a large number of personas is labor-intensive. We propose a method that automatically generates personas for user simulators using a large language model. Furthermore, by assigning personality traits related to communication styles when generating personas, we aim to increase the diversity of communication styles in the user simulator. Experimental results show that the proposed method enables the user simulator to generate utterances with greater variation.
\end{abstract}

\section{Introduction}

Interview dialogue systems, which can efficiently gather information from many users, are a promising application of dialogue system technology and have attracted increasing research interest in recent years \cite{lang2025telephonesurveysmeetconversational,hashimoto-etal-2025-career,zeng-etal-2023-question}.
However, building dialogue systems, including interview dialogue systems, incurs significant costs.
One major cost is the effort required for testing.
The goal of our study is to alleviate the manual effort for testing interview dialogue systems.

To reduce testing costs, {\em user simulation} can be used instead of manual testing.
However, existing user simulators have primarily been developed for the training and evaluation of task-oriented dialogue systems.
Thus, we aim to develop a simulator that is useful for testing interview dialogue systems.

When testing interview dialogue systems, simulating a wide variety of users is crucial.
To increase variation, it is beneficial to have a large number of {\em user personas} \cite{zhang-etal-2018-personalizing,jiang-etal-2024-personallm,hong-etal-2025-dialogue}.
However, manually creating numerous personas is labor-intensive.
While using a large set of pre-constructed personas has been considered \cite{mazare-etal-2018-training}, this approach is unsuitable for interview dialogues, where personas need to be aligned with the specific interview topics.

Therefore, this paper proposes a method to generate diverse personas using {\em large language models} (LLMs).
In this method, a few manually created personas are provided as examples, and the LLM is tasked with generating a large number of additional personas.
To ensure diversity in speaking styles and attitudes toward the system, personality traits are specified during generation.

Evaluation experiments demonstrated that the proposed method enables the creation of diverse dialogues through the generation of many personas.
This increases the likelihood of discovering system issues.
Note that detecting system issues from simulated dialogues is beyond the scope of our study.
Manual inspection is one possible approach, but automated methods using LLMs are also conceivable \cite{finch-etal-2023-leveraging}.

The contributions of our study are as follows:

\begin{itemize}
\item We discuss that diverse personas are necessary for user simulators used in testing interview dialogue systems.
\item We propose a method for generating diverse personas using LLMs.
\item We introduce a technique to enhance persona diversity by specifying personality traits related to communication style.
\item Through evaluation experiments, we demonstrate that the proposed persona generation method enables the creation of diverse dialogues.
\end{itemize}

\section{Related Work}

\subsection{Interview Dialogue Systems}

Interview dialogue systems that extract information from humans through natural language interaction have attracted significant research attention due to their high practical value.
Examples of the application domains include:
course rating surveys at universities \cite{4123399},
telephone surveys \cite{johnston-etal-2013-spoken, lang2025telephonesurveysmeetconversational},
mental health assessments \cite{10.5555/2615731.2617415},
dietary intake recording \cite{kobori-etal-2016-small},
job interviews \cite{su19c_interspeech, b-etal-2020-automatic, inoue:icmi20, JONES2012312, 10.1145/2493432.2493502, 10.5555/2615731.2615838},
dietary preference surveys \cite{zeng-etal-2023-question},
career counseling for nurses \cite{hashimoto-etal-2025-career},
frailty diagnosis \cite{asao-etal-2020-understanding},
collecting people's beliefs about the future of robots \cite{skantze2012furhat},
facilitating user review writing \cite{tanaka-inaba-2024-user}, 
and customer interviews \cite{sidaoui2020ai}.
We propose a method for efficiently testing such interview dialogue systems.

\subsection{Testing Dialogue Systems}

As mentioned earlier, testing dialogue systems requires significant effort, and various methods and tools have been proposed to facilitate this process \cite{li-review}.
For example, there are tools that verify whether a dialogue system behaves as expected using corpora \cite{lintest}, and methods that test the system using predefined test cases \cite{atefi2019automatedtestingframeworkconversational, guo2024mortarmetamorphicmultiturntesting, mutation-test}.
Since these testing methods require the preparation of test cases in advance, they are effective for detecting issues within the expected scope.
However, in interactions with diverse users, unexpected problems may arise, and thus, methods capable of detecting such unforeseen issues are desired.

\subsection{User Simulation for Dialogue Systems}

We therefore use user simulators. User simulators for dialogue systems have traditionally been used to evaluate dialogue strategies \cite{eckert:asru97,niimi:euro99}.  
However, with advances in statistical modeling of dialogue strategies, research has increasingly focused on training dialogue strategy models using reinforcement learning \cite{Schatzmann:survey, Pietquin2016, schatzmann-etal-2007-agenda, ai-litman-2009-setting, li2017usersimulatortaskcompletiondialogues, pietquin2013survey}.

Recently, with the development of large language models (LLMs), user simulators based on LLMs have been proposed \cite{terragni2023incontextlearningusersimulators, luo-etal-2024-duetsim, ALGHERAIRY2025101697, sekulic-etal-2024-reliable, Hu23-Unlocking, Sun23-Metaphorical, di2024linguistics, liu-etal-2023-one}.

In studies on user simulators for task-oriented dialogue systems, evaluation metrics such as similarity to human user behavior and the quality of the learned dialogue strategies have been used \cite{pietquin2013survey}.

Our study differs from these previous studies in that it proposes a method for constructing a user simulator specifically for testing interview dialogue systems. Different evaluation metrics are also required.

\subsection{Using Personas in User Simulation}

Research has also been conducted on using personas in user simulators. \citet{georgila-etal-2008-simulating} and \citet{georgila-etal-2010-learning} demonstrated that different dialogue strategies can be learned by preparing simulators for older and younger users. \citet{hashimoto-etal-2025-career} manually created nurse personas for a user simulator used in evaluating a career counseling dialogue system for nurses, under the supervision of nursing administrators. Our study differs from these studies in that it automatically generates personas.

\subsection{Persona Generation}

As for research on automatic persona generation, there are methods that extract personas from Reddit \cite{mazare-etal-2018-training}. They collect sentences that express a persona (e.g., those containing "I" or "my") and use them for persona creation. Methods for generating personas using LLMs have also been proposed. \citet{ge2024scalingsyntheticdatacreation} proposed a method for generating persona-representing passages from text using an LLM. However, with these methods, it is difficult to generate interviewee personas tailored to the domain of interview dialogue systems.

\section{Proposed Method}

\begin{figure}[t]
\begin{center}
\includegraphics[width=0.5\textwidth]{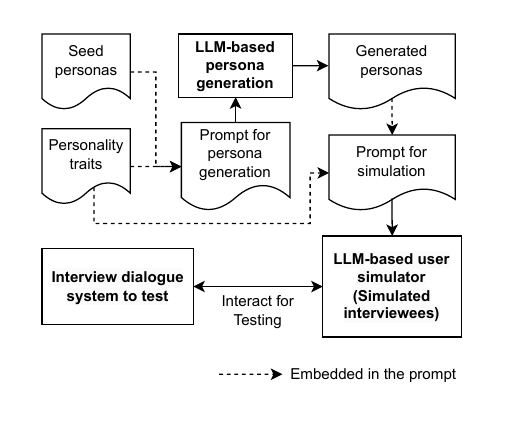}
\end{center}
\caption{Overview of the proposed method.}
\label{method}
\end{figure}

In the proposed method, we generate simulated user personas using an LLM. It is necessary to ensure that the generated personas include information relevant to the domain of the target interview dialogue system. To achieve this, we perform in-context learning using a small number of manually written personas (called {\em seed personas} hereafter) as few-shot examples. Furthermore, to facilitate the discovery of potential issues by introducing greater variation in user utterances during simulation, we provide personality traits that define communication styles when generating personas with the LLM. By embedding the generated personas and personality traits into the prompt of an LLM-based user simulator, we can generate diverse dialogues.
Figure~\ref{method} illustrates the proposed method.

We use the following two axes as personality traits.

\paragraph{Degree of Anthropomorphism}  
One axis concerns whether the user treats the system as an object or as a human. Users who treat the system as an object aim to use the dialogue system efficiently and tend to use expressions that are easy for the system to understand. In contrast, users who treat the system as a human speak to it as if they were speaking to a person, without considering whether the system can understand them. This behavior can be interpreted as users anthropomorphizing the system \cite{reeves1996media}. We refer to this axis as the degree of anthropomorphism.

\paragraph{Degree of Elaborateness}  
The other axis concerns whether the user engages in redundant communication or direct communication. \citet{Pragst2019} and \citet{miehle-etal-2020-estimating} classify users' communication styles into elaborateness and directness. We adopt the same classification in our study.

\begin{figure}[t]
\begin{small}
\begin{tabbing}
- Whether the user often travels\\
- Places the user has recently visited\\
- Favorite places among the user's past travel destinations\\
- Activities the user enjoys doing while traveling\\
- Reasons for not traveling often (if applicable)\\
- Places the user would like to visit next\\
- \= How the user books trips (e.g., visiting a travel agency, \\
  \> calling a travel agency, using a website)
\end{tabbing}
\end{small}
\caption{Topics the travel interview dialogue system asks the user about.}
\label{inteview-content}
\end{figure}

\begin{figure}[t]
\begin{mdframed}[linewidth=1pt]
\begin{footnotesize}
\begin{verbatim}
- Name: Risako Machiyama
- 49 years old
- Female
- Lives in Tokyo
- Lives with husband and two daughters (a senior 
  high school second-year student and a junior 
  high school third-year student)
- Works part-time in an office
- Speaks cheerfully and talks rapidly with 
  many words
- Travels about once a year
- Most recently visited Taiwan with her family
- Enjoys shopping while traveling
- Would like to visit Los Angeles
- Usually books trips by visiting a travel 
  agency
\end{verbatim}
\end{footnotesize}
\end{mdframed}
\begin{mdframed}[linewidth=1pt]
\begin{footnotesize}
\begin{verbatim}
- Name: Yuma Yamanaka
- 32 years old
- Male
- Lives in Hokkaido
- Lives alone
- Chef
- Speaks clearly and often talks about himself
- Travels about once a year
- Most recently visited Osaka alone
- Enjoys food tours during travel for 
  professional inspiration
- Would like to visit Singapore
- Usually books trips online
\end{verbatim}
\end{footnotesize}
\end{mdframed}
\caption{Examples of the seed personas for the travel interview dialogue system.}
\label{seed-personas}
\end{figure}

\begin{figure}

\begin{mdframed}[linewidth=1pt]
\begin{footnotesize}
\begin{verbatim}
# Task

- Create 25 personas for interviewees being
  surveyed about their travel experiences
  and desired travel destinations, generating
  them in JSON format as shown in the example.
- Ensure as much variation among the personas
  as possible.
- The interviewees have the following
  personality traits; create personas
  consistent with these traits.
- Do not reuse any of the personas shown
  in the example.

# Personality Traits

{personality}

# Examples

{examples}
\end{verbatim}
\end{footnotesize}
\end{mdframed}
\caption{Prompt template for persona generation. {\tt \{personality\}} is replaced with a description of personality traits, and {\tt \{examples\}} is replaced with the seed personas converted into JSON format. In the noPT condition, the template without the personality trait description is used.}
\label{persona-genearation-prompt}
\end{figure}

\begin{figure}[t]
APM: High
\begin{mdframed}[linewidth=1pt]
\begin{footnotesize}
\begin{verbatim}
- Views the system as a human rather than an 
  object
- Speaks to the system as if speaking to a person
- Tries to test the system, abruptly changing 
  topics or speaking at length
\end{verbatim}
\end{footnotesize}
\end{mdframed}
APM: Low
\begin{mdframed}[linewidth=1pt]
\begin{footnotesize}
\begin{verbatim}
- Views the system as an object rather than a 
  human
- Aware of the system's limitations in 
  understanding, speaks as simply as possible
- Shows a cooperative attitude toward the system
\end{verbatim}
\end{footnotesize}
\end{mdframed}

\caption{Personality trait descriptions used in the prompt for persona generation by varying the degree of anthropomorphism.}
\label{anthropomorphism}
\end{figure}

\begin{figure}[t]
EL: High
\begin{mdframed}[linewidth=1pt]
\begin{footnotesize}
\begin{verbatim}
- Speaks redundantly
- Talks about things not explicitly asked
- Sometimes uses indirect expressions
\end{verbatim}
\end{footnotesize}
\end{mdframed}
EL: Low
\begin{mdframed}[linewidth=1pt]
\begin{footnotesize}
\begin{verbatim}
- Speaks directly
- Speaks concisely
- Does not talk about unasked topics
\end{verbatim}
\end{footnotesize}
\end{mdframed}
\caption{Personality trait descriptions used in the prompt for persona generation by varying the degree of elaborateness.}
\label{communication-style}
\end{figure}

Although the Big Five personality traits are well-known, we did not use them because it is considered more effective to directly specify users' speaking styles is considered more effective for our purpose.

\section{Evaluation}

We conducted an evaluation experiment to investigate whether the proposed method is effective for testing interview dialogue systems. Note that this experiment deals only with Japanese systems, and that all subsequent figures are translations from Japanese.

\subsection{Compared Methods}

We compared the following five conditions:

\paragraph{BL} (BaseLine):  Only seed personas are used without generating new personas.
\paragraph{noPT} (no Personality Traits):  Proposed method without prompts concerning the two kinds of personality traits. Only seed personas are given when personas are generated.
\paragraph{APM} (AnthroPoMorphism): Proposed method with degree of anthropomorphism given as a personality trait for persona generation. The degree is either High or Low.
\paragraph{EL} (ELaborateness): Proposed method with degree of elaborateness given as a personality trait for persona generation. The degree is either High or Low.
\paragraph{APM+EL} Proposed method with both degree of anthropomorphism and degree of elaborateness given as personality traits for persona generation.

\subsection{Interview Dialogue Systems Used}
\label{systems}

We used the following two systems:

The first is a Japanese text-based interview dialogue system that conducts interviews about travel. It was built using the ChatGPT dialogue built-in block of DialBB \cite{nakano-komatani-2024-dialbb}, where dialogues are conducted based on a single prompt template. The interview topics extracted from users are shown in Figure~\ref{inteview-content}.

The second is a Japanese text-based interview dialogue system that conducts interviews focusing on the user’s preferences for sweets. It asks users questions such as whether they often eat sweets, what kinds of sweets they like, and where they usually buy sweets. This system was also built using DialBB, employing ChatGPT for language understanding, named entity extraction, and dialogue management using a state transition network. ChatGPT is used for evaluating transition conditions and for utterance generation within dialogue management. The state transition network consists of 32 states and 54 transitions.

\begin{figure}[h]
\begin{mdframed}[linewidth=1pt]
\begin{footnotesize}
\begin{verbatim}
# Task Description

- You are acting as a user of a dialogue system 
  and are being interviewed by the system. Based 
  on the flow of the conversation so far, 
  generate your next utterance.

# Notes

- The utterance should consist of either 1, 2, 
  or 3 sentences.
- Do not produce utterances of the same length 
  consecutively.
- Each utterance must be within 100 characters.
- Speak in accordance with your assigned 
  persona below.
- Do not include your name or the word "User" 
  at the beginning of the utterance.
- Answer if you are asked for your name.
- Do not enclose the utterance in quotation 
  marks.

# Your Persona

{persona}

# Dialogue History

{dialogue_history}
\end{verbatim}
\end{footnotesize}
\end{mdframed}
\caption{Prompt template used by the simulator. {\tt \{persona\}} is replaced with the generated persona, and {\tt \{dialogue\_history\}} is replaced with the dialogue history at runtime.}
\label{simulation-prompt}
\end{figure}

\subsection{Procedure}
\label{procedure}

We conducted persona generation and simulation according to the following procedure.

First, for each system, we manually created 10 seed personas. Figure~\ref{seed-personas} shows examples of seed personas for the travel system. In this paper, we show only examples and prompt templates for the travel interview system for the lack of space. 
In the noPT, APM, EL, and APM+EL conditions, these seed personas were embedded into the prompt as few-shot examples to generate new personas.  
In the APM, EL, and APM+EL conditions, personality traits were assigned in a balanced manner, and personas consistent with those traits were generated.  
The template used for persona generation for the travel system is shown in Figure~\ref{persona-genearation-prompt}, and the descriptions of the personality traits used in the experiments are shown in Figures~\ref{anthropomorphism} and \ref{communication-style}.
The third item under "APM: High" in Figure~\ref{anthropomorphism} indicates that users who anthropomorphize the system attempt to test whether the system can engage in conversation as flexibly as a human.
For each condition, 100 personas were generated.\footnote{Personas were generated in batches of about 25 at a time, and when the total exceeded 100, 100 personas were randomly selected.}

The interview dialogue systems interacted with user simulators based on the generated personas.  
The prompt template used for simulation is shown in Figure~\ref{simulation-prompt}.  
Each dialogue session consisted of 15 user utterances.  For conditions other than BL, 
each persona was used for only one dialogue session, resulting in 100 dialogues per condition.
For BL, each of the 10 seed personas was used 10 times.

The LLM used by the persona generator and the simulator was OpenAI's gpt-4o-2024-11-20,\footnote{\url{https://platform.openai.com/docs/models/gpt-4o}} while the
LLM used by the interview dialogue systems was OpenAI's gpt-4o-mini-2024-07-18.\footnote{\url{https://platform.openai.com/docs/models/gpt-4o-mini}} Since our objective is to evaluate the simulator, we used a high-performance model for the simulator. The temperature parameter was set to 0.7 for all LLM usages.

Prompt tuning was conducted separately from the main experiments, using a different system (a sweets interview dialogue system that uses a single prompt template) and different seed personas from those used in the main evaluation described in Section~\ref{systems}.

\begin{table*}[h]
\begin{footnotesize}
\begin{center}

\begin{tabular}{ll|rrrrrrr}
\hline\hline
Condition & Personality & \#Dialogues & \multicolumn{2}{l}{\begin{tabular}[c]{@{}c@{}}Ave. utterance \\ length (S.D.)\end{tabular}} & \begin{tabular}[c]{@{}c@{}}Total \\ words\end{tabular} & \begin{tabular}[c]{@{}c@{}}Unique \\ words\end{tabular} & \begin{tabular}[c]{@{}c@{}}Unique \\ bigrams\end{tabular} & TTR \\ \hline\hline
BL &  & 100 & 28.2 & (7.7) & 42,260 & 15,747 & 31,620 & .373 \\\hline
noPT &  & 100 & 28.5 & ({\bf 7.0}) & 42,784 & 16,362 & 32,730 & .382 \\\hline
APM & All & 100 & 28.4 & (7.2) & 42,563 & 16,089 & 32,334 & .378 \\
 & High & 50 & 30.1 & (7.6) & 22,547 & 8,406 & 17,098 & .373 \\
 & Low & 50 & 26.7 & (6.3) & 20,016 & 7,683 & 15,236 & .384 \\\hline
EL & All & 100 & 36.0 & ({\bf 18.2}) & 53,951 & 18,556 & 39,010 & .344 \\
 & High & 50 & 50.7 & (13.9) & 38,001 & 12,175 & 26,883 & .320 \\
 & Low & 50 & 21.3 & (5.7) & 15,950 & 6,381 & 12,127 & .400 \\\hline
APM+EL & All & 100 & 31.4 & (13.4) & 47,111 & 16,856 & 34,839 & .358 \\
 & High+High & 25 & 46.2 & (12.3) & 17,308 & 5,624 & 12,309 & .325 \\
 & High+Low & 25 & 23.6 & (4.8) & 8,836 & 3,451 & 6,787 & .391 \\
 & Low+High & 25 & 35.2 & (9.9) & 13,184 & 4,651 & 9,796 & .353 \\
 & Low+Low & 25 & 20.8 & (5.9) & 7,783 & 3,130 & 5,947 & .402 \\ \hline
\end{tabular}
~\\~\\~\\

\begin{tabular}{ll|rrrrrrrr}
\hline\hline
Condition & Personality & \begin{tabular}[c]{@{}c@{}}Unique \\ CW\end{tabular} & CW-TTR & SE & CE & MTLD & MSTTR & \multicolumn{2}{l}{\begin{tabular}[c]{@{}c@{}}Ave. TTR \\ (S.D.)\end{tabular}} \\ \hline\hline
BL &  & 1,917 & {\bf .106} & 7.52 & 3.44 & 49.6 & 0.719 & .373 & (.023) \\\hline
noPT &  & 2,282 & {\bf .122} & 7.62 & 3.46 & 51.4 & 0.725 & .384 & (.023) \\\hline
APM & All & 2,050 & .112 & 7.59 & 3.47 & 50.4 & 0.725 & .379 & (.020) \\
 & High & 1,543 & .161 & 7.58 & 3.37 & 52.6 & 0.733 & .374 & (.020) \\
 & Low & 1,418 & .162 & 7.46 & 3.20 & 48.9 & 0.719 & .385 & (.020) \\\hline
EL & All & 2,428 & .104 & 7.64 & 3.62 & 51.7 & 0.731 & .362 & (.046) \\
 & High & 2,127 & .131 & 7.67 & 3.61 & 56.5 & 0.746 & .323 & (.024) \\
 & Low & 1,204 & .171 & 7.27 & 2.96 & 42.8 & 0.696 & .401 & (.024) \\\hline
APM+EL & All & 2,257 & .112 & 7.62 & 3.60 & 50.4 & 0.727 & .369 & (.037) \\
 & High+High & 1,440 & .198 & 7.59 & 3.42 & 54.6 & 0.741 & .327 & (.020) \\
 & High+Low & 908 & .242 & 7.38 & 3.00 & 48.3 & 0.720 & .391 & (.021) \\
 & Low+High & 1,243 & .220 & 7.50 & 3.26 & 51.8 & 0.731 & .356 & (.026) \\
 & Low+Low & 803 & .235 & 7.15 & 2.74 & 43.4 & 0.698 & .403 & (.024) \\ \hline\hline
\end{tabular}
\end{center}
\end{footnotesize}
\caption{Diversity metrics of the simulations for the travel interview system. For example, the row where Condition is APM and Personality is All represents metrics calculated from all the dialogues under the APM condition. The subsequent row, where Personality is High, shows metrics calculated from the dialogues under the APM condition using a high degree of anthropomorphism in the personality setting. 
The row where Condition is APM+EL and Personality is High+High
represents metrics calculated from the dialogues with a high degree of anthropomorphism and a high degree of elaborateness. Bold numbers are mentioned in the main text.}
\label{results1}
\end{table*}

\begin{table*}[h]
\begin{footnotesize}
\begin{center}

\begin{tabular}{ll|rrrrrrr}
\hline\hline
Condition & Personality & \multicolumn{1}{l}{\#Dialogues} & \multicolumn{2}{l}{\begin{tabular}[c]{@{}l@{}}Ave. utterance \\ length (S.D.)\end{tabular}} & \begin{tabular}[c]{@{}c@{}}Total \\ words\end{tabular} & \begin{tabular}[c]{@{}c@{}}Unique \\ words\end{tabular} & \begin{tabular}[c]{@{}c@{}}Unique \\ bigrams\end{tabular} & \multicolumn{1}{l}{TTR} \\ \hline\hline
BL &  & 100 & 25.9 & (7.8) & 25,152 & 11,138 & 20,534 & .443 \\\hline
noPT &  & 100 & 25.6 & ({\bf 8.0}) & 25,509 & 11,105 & 20,634 & .435 \\\hline
APM & All & 100 & 25.6 & (8.2) & 25,680 & 11,124 & 20,862 & .433 \\
 & High & 50 & 26.5 & (8.2) & 13,269 & 5,723 & 10,824 & .431 \\
 & Low & 50 & 24.6 & (8.0) & 12,411 & 5,401 & 10,038 & .435 \\\hline
EL & All & 100 & 33.9 & ({\bf 17.7}) & 34,020 & 13,214 & 26,117 & .388 \\
 & High & 50 & 47.5 & (14.8) & 23,755 & 8,578 & 17,808 & .361 \\
 & Low & 50 & 20.3 & (6.1) & 10,265 & 4,636 & 8,309 & .452 \\\hline
APM+EL & All & 100 & 28.1 & (12.2) & 28,067 & 11,549 & 22,182 & .411 \\
 & High+High & 25 & 39.4 & (13.1) & 9,857 & 3,743 & 7,643 & .380 \\
 & High+Low & 25 & 22.2 & (6.7) & 5,560 & 2,458 & 4,533 & .442 \\
 & Low+High & 25 & 30.6 & (10.5) & 7,642 & 3,087 & 6,015 & .404 \\
 & Low+Low & 25 & 20.0 & (5.9) & 5,008 & 2,261 & 3,991 & .451 \\ \hline\hline
\end{tabular}
~\\~\\~\\
\begin{tabular}{ll|rrrrrrrr}
\hline\hline
Condition & Personality & \multicolumn{1}{l}{\begin{tabular}[c]{@{}c@{}}Unique \\ CW\end{tabular}} & \multicolumn{1}{l}{CW-TTR} & \multicolumn{1}{l}{SE} & \multicolumn{1}{l}{CE} & \multicolumn{1}{l}{MTLD} & \multicolumn{1}{l}{MSTTR} & \multicolumn{2}{l}{\begin{tabular}[c]{@{}c@{}}Ave. TTR\\ (S.D.)\end{tabular}} \\ \hline\hline
BL &  & 1,157 & {\bf .109} & 7.35 & 3.09 & 53.4 & 0.732 & .444 & (.023) \\\hline
noPT &  & 1,458 & {\bf .133} & 7.43 & 3.18 & 53.9 & 0.736 & .437 & (.029) \\\hline
APM & All & 1,547 & .141 & 7.51 & 3.33 & 53.2 & 0.737 & .435 & (.024) \\
 & High & 1,090 & .193 & 7.45 & 3.15 & 56.6 & 0.740 & .432 & (.020) \\
 & Low & 1,081 & .202 & 7.41 & 3.11 & 51.3 & 0.724 & .438 & (.028) \\\hline
EL & All & 1,838 & .127 & 7.58 & 3.45 & 54.2 & 0.740 & .409 & (.053) \\
 & High & 1,553 & .156 & 7.58 & 3.41 & 57.2 & 0.753 & .364 & (.027) \\
 & Low & 933 & .208 & 7.24 & 2.90 & 47.3 & 0.715 & .454 & (.028) \\\hline
APM+EL & All & 1,604 & .134 & 7.51 & 3.38 & 52.7 & 0.731 & .421 & (.038) \\
 & High+High & 981 & .240 & 7.49 & 3.16 & 56.5 & 0.750 & .383 & (.026) \\
 & High+Low & 649 & .272 & 7.20 & 2.85 & 49.8 & 0.729 & .443 & (.025) \\
 & Low+High & 808 & .246 & 7.33 & 2.94 & 52.6 & 0.726 & .407 & (.032) \\
 & Low+Low & 628 & .284 & 7.08 & 2.62 & 47.0 & 0.709 & .452 & (.024) \\ \hline\hline
\end{tabular}

\end{center}
\end{footnotesize}
\caption{Diversity metrics of the simulations for the sweets interview system.}
\label{results2}
\end{table*}

\subsection{Evaluation Metrics}

Our goal is to automatically expose issues in interview dialogue systems.  
One direct way to evaluate the proposed method would be to build a faulty interview dialogue system, interact with it using the user simulator generated by the proposed method, and measure how many issues can be detected.  
However, faulty systems may cause dialogues to collapse once a problem occurs.  
In practice, using the user simulator to improve a system would involve repeatedly fixing detected issues and running new simulations, but evaluating this iterative process is impractical.

Therefore, as a second-best approach, we measure the diversity of user simulator utterances.  
The rationale is that the more diverse the utterances, the higher the probability of uncovering system issues.

Following previous studies on user simulators for task-oriented dialogue systems \cite{terragni2023incontextlearningusersimulators,ALGHERAIRY2025101697,sekulic-etal-2024-reliable}, we used the following metrics to evaluate utterance diversity.
For Japanese word tokenization, we used Sudachi \cite{TAKAOKA18.8884} in C mode (a mode that does not split compound words).

\paragraph{Average and S.D. of utterance lengths} The average and standard deviation of the number of words per user utterance across all dialogues.

\paragraph{Total words} The total number of words across all user utterances in all dialogues.

\paragraph{Unique words} The number of unique words across all user utterances in all dialogues.

\paragraph{Unique bigrams} The number of unique bigrams across all user utterances in all dialogues.

\paragraph{TTR} Type-Token Ratio:  ((\# of unique words) / (\# of total words)) for all user utterances in all dialogues.

\paragraph{Unique CW} The number of Unique Content Words appearing in all user utterances.

\paragraph{CW-TTR} The type-token ratio for content words across all user utterances.

\paragraph{SE} Shannon Entropy calculated from all user utterances across all dialogues.

\paragraph{CE} Conditional bigram Entropy calculated from all user utterances across all dialogues.

\paragraph{MTLD} Measure of Textual Lexical Diversity \cite{mccarthy2010mtld} calculated by concatenating all user utterances (threshold: 0.72).

\paragraph{MSTTR} Mean Segmental Type-Token Ratio \cite{mccarthy2010mtld} calculated by concatenating all user utterances.

\paragraph{Average and S.D. of TTR} The average and standard deviation of type-token ratios calculated per dialogue.

\subsection{Results}

Tables~\ref{results1} and \ref{results2} respectively show the diversity metrics for the travel interview system and the sweets interview system.  
Results for each personality trait condition are also presented.
These results suggest the following two points:  
(1) noPT exhibits greater content variation compared to BL. In other words, persona generation using LLMs can increase the variety of content in utterances.  
(2) EL shows greater stylistic variation than noPT. That is, by varying redundancy levels, one can increase stylistic diversity in utterances.

(1) is suggested by the following findings: Comparing BL and noPT, especially in the travel domain, CW-TTR increases from $.106$ to $.122$ in the travel domain, and in the sweets domain, from $.109$ to $.133$.
This suggests that the generated personas contain different content words from the seed personas. This may help uncover issues caused by diverse utterance content.
Although SE, CE, and MTLD also show slight improvements, the differences are not significant.
TTR slightly decreases in the sweets domain system, possibly due to limited stylistic changes, leading to low variation in function words.

(2) is suggested by the following findings:  
Compared to noPT, EL (All) shows an increase in the standard deviation of utterance length—from $7.0$ to $18.2$ in the travel domain and from $8.0$ to $17.7$ in the sweets domain.
This suggests that varying the degrees of elaborateness enables the generation of utterances with diverse lengths, potentially exposing problems caused by such diversity.
TTR and CW-TTR are lower, likely because elaborate utterances under the EL High condition tend to include fixed phrases.

There is little difference between APM and noPT. Also, comparing APM Low and High settings reveals no significant variation.
Furthermore, APM+EL does not differ much from EL. Even when examining the dialogues, major differences were not observed.
This suggests that the personality traits described in APM may not be sufficient to alter dialogue style.
Since personas generated in noPT already contain variation in personality, APM might not have introduced additional diversity beyond that.

Note that in settings like EL High, where personality traits are narrowly defined, TTR and CW-TTR increase.
This is likely because the number of dialogues is small, reducing word repetition.

\begin{figure}[t]
noPT:
\begin{mdframed}[linewidth=1pt]
\begin{footnotesize}
\begin{verbatim}
- Name: Erina Takahashi
- 29 years old
- Female
- Lives in Kyoto Prefecture
- Lives alone
- Works for an IT company
- Reserved, but talkative with close friends
- Takes several trips a year
- Most recently visited Kanazawa; traveled
  alone
- Enjoys visiting local art museums and cafes
  while traveling
- Would like to visit Paris, France
- Books flights and hotels online
\end{verbatim}
\end{footnotesize}
\end{mdframed}

\caption{Examples of generated personas (1).}
\label{generated_personas1}
\end{figure}

\begin{figure}[h]
EL High:
\begin{mdframed}[linewidth=1pt]
\begin{footnotesize}
\begin{verbatim}
- Name: Ken Nakamura
- 34 years old
- Male
- Live in Hiroshima Prefecture
- Lives with his wife and 2-year-old son
- Public servant
- Speaks gently and is family-oriented
- Travels less frequently since having a child,
  but still goes on one trip per year
- Most recently visited Kagawa; a family trip
  focused on touring udon restaurants
- Chooses travel destinations that can be
  enjoyed by the whole family
- Would like to visit Hokkaido
- Plans trips together with his wife
\end{verbatim}
\end{footnotesize}
\end{mdframed}

EL Low:
\begin{mdframed}[linewidth=1pt]
\begin{footnotesize}
\begin{verbatim}
- Name: Yusuke Ogawa
- 45 years old
- Male
- Lives in Hiroshima Prefecture
- Lives with his wife and two children (a son
  in 8th grade and a daughter in 5th grade)
- Public servant
- Speaks calmly and avoids unnecessary talk
- Goes on one trip per year
- Most recently visited Kyoto on a family trip
- Visits temples and shrines during travels
- Would like to visit Nara
- Makes travel reservations by calling a travel
  agency
\end{verbatim}
\end{footnotesize}
\end{mdframed}

\caption{Examples of generated personas (2).}
\label{generated_personas2}
\end{figure}

\begin{figure}[h]
APM High:
\begin{mdframed}[linewidth=1pt]
\begin{footnotesize}
\begin{verbatim}
- Name: Kenta Omura
- 24 years old
- Male
- Lives in Osaka Prefecture
- Lives alone
- Graduate student
- Quiet and not very talkative, but polite
- Takes about one trip per year
- Most recently visited Nara; did some 
  sightseeing after attending an academic 
  conference
- Enjoys exploring nature at travel
  destinations
- Would like to visit the Rocky Mountains in
  Canada
- Books flights online by himself
\end{verbatim}
\end{footnotesize}
\end{mdframed}

APM Low:
\begin{mdframed}[linewidth=1pt]
\begin{footnotesize}
\begin{verbatim}
- Name: Naoto Yamashita
- 33 years old
- Male
- Lives in Hiroshima Prefecture
- Lives alone
- Designer
- Speaks calmly and politely
- Takes two trips a year
- Most recently visited Kyoto alone
- Enjoys visiting architectural sites and art
  museums while traveling
- Would like to visit Paris
- Plans and books trips meticulously online
\end{verbatim}
\end{footnotesize}
\end{mdframed}

\caption{Examples of generated personas (3).}
\label{generated_personas3}
\end{figure}

\subsection{Generated Personas}

Examples of the generated personas are shown in Figures~\ref{generated_personas1}, \ref{generated_personas2}, and \ref{generated_personas3}. 
These examples were randomly selected from the personas generated under each condition.
Figures~\ref{generated_personas2} and \ref{generated_personas3} feature only male personas, but this is purely coincidental.
While the male personas shown speak in a reserved manner, some of the other generated male personas are sociable and talkative.
From these examples, it can be inferred that the generated personas are consistent with their respective personality traits.
The APM+EL examples are omitted, as they did not yield any notable results with respect to the diversity metrics.
Additional examples of generated personas are shown in Figures~\ref{addtional_personas1} and \ref{addtional_personas2} in the Appendix.

\section{Discussion}
\label{discussion}

As discussed above, the experimental results suggest that generating personas with an LLM enables the generation of dialogues with greater variation. Although the quantitative metrics do not indicate this very clearly, you can see the variation by looking at the generated dialogues.
This could lead to the discovery of more problems in dialogue systems.  
Furthermore, specifying personality traits related to communication style can introduce greater variation in utterance lengths, which may help uncover additional problems.

In the current user simulator prompt we used, relatively similar utterances tend to be generated consecutively. This might be different from the behaviors of human users.
It is possible that generating various types of utterances within a single dialogue could help reveal different types of issues.  
In future work, we will explore methods for increasing such variation and investigate to what extent the behaviors of the user simulator cover those of human users.

Although our study focused on APM and EL traits, there may be other personality traits that could introduce further variation.  
Identifying and incorporating such traits will be considered in future research.

The ultimate goal of our study is to identify as many issues in interview dialogue systems as possible. However, in this work, we only measured the diversity of simulated users. Whether the proposed method can help discover issues across various dialogue systems remains a topic for future work. In the future, we aim to integrate this method into tools for building interview dialogue systems and validate it through its application in the development of diverse systems.

Note that, although the systems used in our evaluation experiments did not exhibit obvious issues such as completely incoherent utterances, we found that under the El High condition, simulated users tended to engage in extended small talk.  This often caused the system to fail in extracting the necessary information, revealing a potential issue.

\section{Concluding Remarks}

This paper proposed a persona generation method for user simulation using large language models (LLMs). We also introduced a method to incorporate personality traits related to communication style during persona generation. This enables diverse testing of interview dialogue systems without human involvement, significantly reducing development costs. Although the evaluation experiments were limited and some issues remain, the results suggest the effectiveness of the proposed method. Therefore, we believe it is worthwhile to share our proposed method and experimental findings.

In the future, we plan to develop user simulators for interview dialogue systems with speech input/output and multimodal input/output.  
We also aim to extend our approach to build user simulators for various types of dialogue systems beyond interview dialogues.

\section*{Acknowledgments}

This work was partially inspired by the collaboration between Ekai Hashimoto, Takayoshi Sakurai, and Shun Shiramatsu of Nagoya Institute of Technology, Toshitake Komazaki of Tokyo Healthcare University, Shiho Tsuchiya of Kitasato University Hospital, and the first author. We sincerely thank them.

\section*{Limitations}

In addition to what was discussed in Section \ref{discussion}, there are several other limitations in our evaluation experiments. First, we used only a single model as the LLM for both persona generation and the simulator. Second, we used only Japanese systems. Furthermore, the number of generated personas per condition was fixed at 100, the number of seed personas was set to 10 for both systems, and the temperature parameter was set to 0.7 throughout the experiment; other configurations have not been tested. The impact of varying these conditions remains unexamined.

In future work, we plan to address these issues as we apply the proposed method to the development of various interview dialogue systems.

\section*{Ethical Considerations}

In our study, human participants were not involved in system testing; therefore, there is no risk of collecting or leaking personal information from individuals.

The biases potentially inherent in LLMs \cite{10.1162/coli_a_00524} may be reflected in automatically generated personas or user utterances. As a result, certain types of personas or utterances may be excluded from testing, which could hinder the identification of potential issues. Verifying whether this problem occurs and addressing it if necessary remain future challenges.

\bibliography{custom}

\appendix

\section{Example Dialogues from the Travel Interview System}
\label{sec:appendix}

Example dialogues from the travel interview system are shown in Figures~\ref{dialogue_example1}--\ref{dialogue_example4}.  
Figure~\ref{dialogue_example1} shows a result from testing with a simulator using the lower persona from Figure~\ref{seed-personas}.  
Figures~\ref{dialogue_example2}--\ref{dialogue_example4} show results from testing with simulators using the generated personas in Figures~\ref{generated_personas1}, \ref{generated_personas2}, and \ref{generated_personas3}.

\begin{figure*}[t]
\begin{center}
\begin{footnotesize}
\begin{tabular}{lp{13cm}}
System & Hello, my name is Shota Watanabe. I'd like to interview you today about travel. Do you travel often? \\
User & About once a year, I'd say. I recently went to Osaka. I strolled around on my own and enjoyed trying out different foods. \\
System & Eating your way through Osaka sounds fun! Was there any dish that left a strong impression on you? \\
User & Definitely the takoyaki. I had some at a local shop—the outside was crispy and the inside was soft and gooey. It was amazing. The kushikatsu was also delicious. \\
System & Takoyaki and kushikatsu both sound tasty! Have there been any other travel destinations that you particularly enjoyed? \\
User & Kyoto was great too. I really liked the historic atmosphere of the streets, and I tried a variety of traditional Japanese sweets. \\
System & The historical streets of Kyoto sound wonderful! Were there any sweets that especially stood out to you? \\
User & The dorayaki left the biggest impression. The fluffy pancake and the mildly sweet bean paste were a perfect match. I also enjoyed yatsuhashi. \\
System & Dorayaki and yatsuhashi—both sound delicious! What do you enjoy most when traveling? \\
User & Food tours are my favorite. I love discovering local flavors unique to each area. I do a bit of sightseeing too. \\
System & So eating local food is the most fun part! Where would you like to go next? \\
User & I'd love to visit Singapore. I want to try chicken rice and hawker stall dishes. I'm also looking forward to experiencing the atmosphere of the city. \\
System & Hawker food in Singapore is definitely appealing! How do you usually make your travel arrangements? \\
User & I usually book everything online. It's convenient to search for flights and hotels all in one place. \\
System & Online booking really is convenient! Which sites do you often use? \\
User & Mostly the major travel websites. They make it easy to compare options, and I always check for campaign deals. I go with the ones I find easiest to use. \\
System & Comparing travel sites is important! How do you keep track of special deals and offers? \\
User & I use site notifications and subscribe to email newsletters. I check them regularly so I don't miss any good deals. \\
System & It's great that you keep on top of special offers! Is there anything else you pay attention to when planning a trip? \\
User & I make a list of places I want to visit, and I try to plan carefully around time and budget. I also check the weather. \\
System & Making a list sounds like a smart idea! Have you started planning your next trip? \\
User & Not in detail yet, but I'm hoping to go to Singapore next year. I'm looking forward to planning it out! \\
System & Sounds exciting! Are there any other countries or regions you'd like to visit? \\
User & I'm also interested in Europe. I'd especially like to try authentic cuisine in Italy and Spain. \\
System & European cuisine is certainly appealing! Are there any specific dishes you'd like to try in Italy or Spain? \\
User & In Italy, I want to try pizza and pasta. In Spain, definitely paella and tapas. I'm especially interested in dishes that use local ingredients. \\
System & Local ingredient-based dishes are wonderful! How do you choose where to eat when you travel? \\
User & I usually look up popular places or restaurants loved by locals. I also rely on reviews and word-of-mouth recommendations. \\
System & Choosing popular local spots sounds like a great strategy! Is there anything else you look forward to when traveling? \\
User & I enjoy experiencing the local scenery and culture too. I love immersing myself in the unique atmosphere of each place. \\
System & Experiencing the scenery and culture is definitely rewarding! Was there a place that left a particularly strong impression on you? \\
\end{tabular}
\end{footnotesize}
\end{center}
\caption{Example simulated dialogue in condition BL.}
\label{dialogue_example1}
\end{figure*}

\begin{figure*}[t]
\begin{center}
\begin{footnotesize}
\begin{tabular}{lp{13cm}}
System & Hello, my name is Shota Watanabe. I'd like to interview you today about travel. Do you travel often? \\
User & I travel a few times a year. I recently took a solo trip to Kanazawa. \\
System & Kanazawa is a lovely place! Was there anything in particular that left an impression? \\
User & Yes, the exhibition I saw at a local art museum was very memorable. I also really enjoyed visiting different cafes. \\
System & Visiting art museums and cafes sounds wonderful! Have there been other destinations you enjoyed? \\
User & Yes, though it's close to Kyoto, I also liked Nara. I appreciated the calm and quiet atmosphere. \\
System & The peaceful atmosphere of Nara is indeed charming! What do you enjoy doing when you travel? \\
User & I like visiting places where I can experience local culture and history. I also enjoy taking photographs. \\
System & Experiencing local culture and history is great! Where would you like to travel next? \\
User & I'd like to visit Paris, France. I want to fully enjoy the museums and the cityscape. \\
System & Paris is a fascinating destination! How do you usually make your travel reservations? \\
User & I usually book flights and hotels online. It's convenient and makes it easy to compare options. \\
System & Online booking is certainly convenient! Which websites do you use most often? \\
User & I usually use major travel booking sites. I appreciate the abundance of user reviews. \\
System & Having plenty of reviews is important! Do you consult any other sources for information? \\
User & I also refer to travel blogs and social media posts. Photos in particular help me get a better sense of what to expect. \\
System & Travel blogs and social media are great sources! Which social media platforms do you use most? \\
User & I mainly use Instagram. There are so many travel photos that really spark my imagination. \\
System & Instagram is perfect for travel inspiration! Are there any other countries you'd like to visit? \\
User & I'd like to visit Italy too. Especially to tour museums in Florence and Rome. \\
System & Touring museums in Italy sounds fantastic! Are there any specific museums you're looking forward to? \\
User & The Uffizi Gallery is at the top of my list. I'd love to take my time viewing Renaissance masterpieces. \\
System & The Uffizi is a great choice! Are there any other artworks you're interested in? \\
User & I'm also interested in Impressionist works by Monet and Renoir. Their use of light is especially beautiful. \\
System & Monet and Renoir's works are lovely indeed! Are there any other artists you like? \\
User & I also like Klimt. The use of gold leaf in ``The Kiss" is particularly striking. \\
System & Klimt's ``The Kiss" is truly a beautiful piece! Are there other museums you'd like to visit? \\
User & I'd like to go to the Orsay Museum as well. They have many Impressionist works, and I'd love to explore them thoroughly. \\
System & The Orsay Museum is also a wonderful choice! Besides museum visits, is there anything else you're looking forward to? \\
User & I'd like to stroll through the streets of Paris at a relaxed pace. I'm also looking forward to spending some leisurely time at cafes. \\
System & Spending time at cafes sounds lovely! Are there any cafes in particular you'd like to visit? \\
\end{tabular}
\end{footnotesize}
\end{center}
\caption{Example simulated dialogue in condition noPT.}
\label{dialogue_example2}
\end{figure*}

\begin{figure*}[t]
\begin{center}
\begin{footnotesize}
\begin{tabular}{lp{13cm}}
System & Hello, my name is Shota Watanabe. I'd like to interview you today about travel. Do you travel often? \\
User & Not so frequently these days since our child is still small, but we try to take a family trip about once a year. \\
System & A yearly family trip sounds wonderful! Where did you go most recently? \\
User & We went to Kagawa recently. We toured several udon restaurants as a family, and our son really seemed to enjoy it. The scenery in Kagawa was beautiful and very relaxing as well. \\
System & A tour of udon restaurants in Kagawa sounds fun! Were there any other places you've enjoyed visiting? \\
User & Yes, a little while ago we went to Kyoto. It has many historical sites, and my wife and I enjoyed strolling around. That was before our son was born, so we were able to take it easy. \\
System & The historical sites in Kyoto are wonderful! What do you find most enjoyable when you travel? \\
User & The most important thing is doing something that the whole family can enjoy together. Lately, we've been choosing destinations and activities our son would like. We enjoy sightseeing, but discovering delicious local food is also a big part of the fun. \\
System & Family-friendly activities and good food—that sounds lovely! Is there somewhere you'd like to go next? \\
User & I'd like to visit Hokkaido. The vast natural landscapes and delicious food are really appealing. I also hope we can experience the snowy scenery as a family someday. \\
System & Hokkaido is a great choice! How do you usually plan your trips? \\
User & I usually plan the trip together with my wife. We research online and check reviews. We try to be thorough so that everyone in the family can enjoy the trip. \\
System & Online research is helpful! Reviews are especially valuable. Which websites do you use? \\
User & We often use travel review sites and accommodation booking sites. We tend to look for sites that have a lot of family-friendly information. We also find blogs and social media posts from people who've actually been there quite helpful. \\
System & Review sites and social media are very useful! Which site has been the most helpful? \\
User & Recently, we've been using Jalan and Rakuten Travel. They offer many family-friendly plans, and the reviews are detailed and informative. Reading about others' experiences really builds anticipation. \\
System & Jalan and Rakuten Travel are great options! Are there any other destinations you're interested in? \\
User & Yes, we're also interested in Okinawa. The ocean is beautiful, and it seems like a great place to relax as a family. Once our son is a bit older, we'd love to try snorkeling too. \\
System & Okinawa sounds like a wonderful choice! Snorkeling sounds fun. Is there a specific beach you'd like to visit? \\
User & We haven't decided on a specific beach yet, but we'd like to go somewhere with clear water. A calm beach where our son can play in the sand would be ideal. My wife and I are planning to look into places that are kid-friendly. \\
System & A calm beach sounds perfect for a relaxing family time! Do you have any other memorable travel experiences? \\
User & I remember when our son took his first bullet train ride on a trip. He was a bit nervous at first, but he really enjoyed looking out the window. We also shared a boxed lunch on the train—it was a fun memory for all of us. \\
System & What a lovely memory on the bullet train! Any other travel stories that left an impression? \\
User & When we did the udon tour in Kagawa, our son said, “I want to eat more!” That really stuck with me. It made us all laugh, and the atmosphere was so warm and cheerful. It's often those small, everyday moments during travel that become the most memorable. \\
System & Those little moments during travel really stay with you! Are there any other countries or regions you'd like to visit? \\
User & I'd love to visit Europe, especially Italy and France. I'm drawn to the beautiful cityscapes and historic architecture. I think it would be great to take a leisurely family trip there. \\
System & Europe, especially Italy and France, sounds wonderful! Do you have any specific cities in mind? \\
User & Yes, in Italy, I'd like to visit Florence and Venice. In France, Paris is definitely a dream destination. Both places are rich in history and culture, and I think walking around as a family would be really enjoyable. \\
System & Florence and Venice are excellent choices! In Paris, is there a particular attraction you're most interested in? \\
User & I'd really like to visit the Eiffel Tower and the Louvre Museum. They're famous spots, but I think they're places you have to see at least once. I also think a stroll along the Seine with my family would be lovely. \\
System & The Eiffel Tower and the Louvre sound amazing! Are there any other activities you'd like to try? \\
\end{tabular}
\end{footnotesize}
\end{center}
\caption{Example simulated dialogue in condition EL High.}
\label{dialogue_example5}
\end{figure*}

\begin{figure*}[t]
\begin{center}
\begin{footnotesize}
\begin{tabular}{lp{13cm}}
System & Hello, my name is Shota Watanabe. I'd like to interview you today about travel. Do you travel often? \\
User & I travel about once a year. \\
System & A yearly trip sounds great! Where did you go most recently? \\
User & I went to Kyoto. It was a family trip where we visited various temples and shrines. \\
System & Visiting temples and shrines in Kyoto sounds wonderful! Was there a place that left a strong impression? \\
User & Kiyomizu-dera Temple stood out the most. The scenery was beautiful, and my whole family enjoyed it. \\
System & The view from Kiyomizu-dera is lovely! Were there any other enjoyable activities during the trip? \\
User & Choosing souvenirs with my children was a lot of fun. My daughter especially seemed very happy. \\
System & Picking out souvenirs is always enjoyable! Where would you like to travel next? \\
User & I'm thinking of visiting Nara. I'd like to tour the historic temples and shrines there. \\
System & Nara is a great destination too! How do you usually make your travel reservations? \\
User & I make reservations by calling a travel agency. I feel more secure speaking directly with someone. \\
System & Speaking directly does provide peace of mind. How is the service at the travel agency? \\
User & They're polite and explain things clearly. They're also very flexible with our requests. \\
System & Good service at a travel agency is important! Are there any other points you prioritize when traveling? \\
User & I focus on making sure the whole family has fun, and that transportation and accommodation are smooth and stress-free. \\
System & Ensuring the whole family enjoys the trip is very important! Do you make any special arrangements for transport or lodging? \\
User & We choose transportation that minimizes stress, and we prefer places where the family can relax comfortably together. \\
System & Those are thoughtful considerations! Do you have any memorable travel experiences? \\
User & I remember how excited my kids were when they rode the bullet train for the first time. \\
System & Their first bullet train ride—that must have been special! Any other memorable travel moments? \\
User & One time we all wore yukata and strolled through the garden of our inn. It felt fresh and different from our everyday life. \\
System & Walking in yukata sounds like a unique experience! Are there other places you'd like to visit? \\
User & I'd like to go to Hokkaido. It seems to have rich nature and many places the whole family can enjoy. \\
System & Hokkaido is a great choice! Are there any activities you're especially looking forward to? \\
User & I'd like to go hiking with my family in nature. I've heard there are many scenic places. \\
System & Hiking sounds fantastic! Are there any foods you'd like to try or experiences you'd like to have? \\
User & I'd like to try fresh seafood--especially crab and sea urchin. \\
System & Fresh seafood is very appealing! Would you like help finding recommended restaurants in Hokkaido? \\
User & Thank you, but I prefer to look for restaurants once I'm there. I only do a little research in advance. \\
System & Exploring on-site sounds fun too! Is there anything in particular you pay attention to when preparing for a trip? \\
\end{tabular}
\end{footnotesize}
\end{center}
\caption{Example simulated dialogue in condition EL Low.}
\label{dialogue_example6}
\end{figure*}

\begin{figure*}[t]
\begin{center}
\begin{footnotesize}
\begin{tabular}{lp{13cm}}
System & Hello, my name is Shota Watanabe. I'd like to interview you today about travel. Do you travel often? \\
User & About once a year. I recently went to Nara. It was for a conference, but I did some sightseeing as well. \\
System & Nara is a historic place! Was there anything that particularly impressed you during your sightseeing? \\
User & Todai-ji Temple left a strong impression. I was overwhelmed by the sheer presence of the Great Buddha. I also found the deer very soothing. \\
System & The Great Buddha at Todai-ji really is impressive! Were there any other enjoyable experiences during your travels? \\
User & I enjoy walking in nature, so strolling through Nara's green parks was a lot of fun. \\
System & Walking in nature is wonderful! Where would you like to go next? \\
User & I'd like to visit the Rocky Mountains in Canada. I want to hike through the vast natural landscapes. \\
System & The Rocky Mountains are a fantastic choice! How do you usually book your trips? \\
User & I usually arrange flights and accommodation online by myself. Which websites do you often use? \\
System & So you typically book online. I often use comparison websites myself. Do you have any recommended sites? \\
User & I don't stick to any particular site, but I often use Skyscanner or Expedia. \\
System & Skyscanner and Expedia are quite handy! Is there anything else you pay attention to when planning a trip? \\
User & I try to plan within my budget and avoid overdoing it. I also make sure to check local transportation in advance. \\
System & Checking your budget and transport options is important! How often do you usually travel? \\
User & About once a year. I can't go that often, so I try to plan thoroughly when I do. \\
System & It's great that you plan thoroughly! Is there a reason you can't travel more often? \\
User & I don't have much free time. My graduate research keeps me pretty busy. \\
System & It sounds like your research takes a lot of your time. It must be hard to secure quality time. How do you plan your trips? \\
User & I plan them when my research schedule is more relaxed. I try to decide on the destination and dates early. \\
System & Planning early is a good idea! Do you already have specific dates in mind for your next trip? \\
User & Not yet, but I'll start thinking about it once things calm down with my research. \\
System & Once your research calms down, how long are you thinking of traveling for? \\
User & I'm thinking about a week. If it's too long, it could affect my research, so a balanced length is ideal. \\
System & A week-long trip sounds great! Do you have a budget in mind? \\
User & I haven't decided exactly, but I'm estimating around 200,000 yen, including flights and accommodation. \\
System & Around 200,000 yen--got it. Do you have any specific activities in mind for the trip? \\
User & I think the main activity will be hiking in nature. I'd also like to try some local food. \\
System & Enjoying local food sounds wonderful! Is there anything specific you'd like to eat? \\
User & Nothing in particular, but I'd like to try local specialties and dishes made with fresh ingredients. \\
System & Enjoying local specialties is a great idea! Is there anything else you look forward to when traveling? \\
\end{tabular}
\end{footnotesize}
\end{center}
\caption{Example simulated dialogue in condition APM High.}
\label{dialogue_example3}
\end{figure*}

\begin{figure*}[t]
\begin{center}
\begin{footnotesize}
\begin{tabular}{lp{13cm}}
System & Hello, my name is Shota Watanabe. I'd like to interview you today about travel. Do you travel often? \\
User & I travel about twice a year. Most recently, I went to Kyoto. \\
System & Kyoto is a wonderful place! Was there anything that left a particular impression on you? \\
User & I toured various architectural sites and museums alone, and the Five-Storied Pagoda at To-ji Temple was especially memorable. \\
System & The Five-Storied Pagoda at To-ji is indeed impressive! Were there any other places you enjoyed visiting? \\
User & I also liked Onomichi, which is relatively close to Hiroshima. The sloping streets and old townscape were very charming. \\
System & The slopes and townscape of Onomichi sound lovely! What do you enjoy doing when you travel? \\
User & I enjoy visiting architectural sites and museums. It's fascinating to experience the history and design. \\
System & Engaging with history and design is definitely appealing! Is there a place you'd like to visit next? \\
User & I'd like to go to Paris. I especially want to see the Louvre Museum and the Eiffel Tower. \\
System & Paris is a wonderful choice! How do you usually make your reservations? \\
User & I research online and make detailed plans before booking. It helps me avoid wasting time. \\
System & So you're a well-organized traveler! Which websites do you usually use? \\
User & Mainly travel booking sites and official museum websites. I also gather information from review sites. \\
System & Sounds like you plan your trips thoroughly! Which review sites have been particularly helpful? \\
User & It depends on the destination, but I often use TripAdvisor. There's a wealth of information there. \\
System & TripAdvisor is definitely useful! Are there any other sites you recommend? \\
User & I also find Expedia convenient and easy to use. I often book flights and accommodations together. \\
System & Expedia is a good choice too! How do you choose your accommodations? \\
User & I prioritize location and cleanliness. I always make sure to read reviews carefully. \\
System & Choosing accommodations wisely is important! What kind of reviews do you find most helpful? \\
User & Reviews that include detailed impressions and photos from guests are especially useful. I also check the average rating. \\
System & Guest impressions are certainly valuable! Do you already have your next destination in mind? \\
User & Not yet, but I'd like to choose a place where I can engage with history and design again. \\
System & A place rich in history and design sounds exciting! Do you have any specific candidates? \\
User & I'm interested in other European cities, such as Florence and Barcelona. \\
System & Florence and Barcelona are both great choices! Why do you want to visit them? \\
User & I want to experience Renaissance art and architecture in Florence. As for Barcelona, Gaud\'{i}'s architecture is very appealing. Both cities seem rich in design and history, which fascinates me. \\
System & Florence and Barcelona are both wonderful! Are there any other countries you'd like to visit? \\
User & I'm also interested in Switzerland. I'd like to see landscapes where nature and architecture blend together. \\
System & Switzerland's fusion of nature and architecture sounds amazing! Is there any particular place you'd like to visit there? \\
\end{tabular}
\end{footnotesize}
\end{center}
\caption{Example simulated dialogue in condition APM Low.}
\label{dialogue_example4}
\end{figure*}

\section{Additional Examples of Generated Personas}
\label{sec:addtional-personas}

Figures~\ref{addtional_personas1} and \ref{addtional_personas2} show additional examples of generated personas. These were also randomly selected.

\begin{figure}[t]
noPT:
\begin{mdframed}[linewidth=1pt]
\begin{footnotesize}
\begin{verbatim}
- Name: Yusuke Matsumoto
- 33 years old
- Male
- Lives in Nagano Prefecture
- Lives with his parents
- Farmer
- Quiet but sincere; good at detailed tasks
- Goes on a trip once a year
- Recently visited Nagoya alone
- Visits markets and direct-from-farm produce 
  stands when traveling
- Would like to visit Spain
- Does his own research and makes reservations 
  online
\end{verbatim}
\end{footnotesize}
\end{mdframed}

EL High:
\begin{mdframed}[linewidth=1pt]
\begin{footnotesize}

\begin{verbatim}
- Name: Kaori Ishikawa
- 48 years old
- Gender: Female
- Lives in Niigata Prefecture
- Lives with her husband and their junior high 
  school-aged daughter
- Nursery school teacher
- Speaks in a gentle tone and is polite in 
  conversation
- Goes on a trip about once a year
- Recently visited Sendai on a family 
  sightseeing trip
- Enjoys soaking in hot springs while traveling
- Would like to visit Kyoto
- Plans trips using travel magazines as a 
  reference
\end{verbatim}
\end{footnotesize}
\end{mdframed}

EL Low:
\begin{mdframed}[linewidth=1pt]
\begin{footnotesize}
\begin{verbatim}
- Name: Misaki Yamauchi  
- 47 years old
- Female  
- Lives in Nagasaki Prefecture  
- Lives with her husband  
- Homemaker
- Speaks in a gentle and calm manner  
- Travels once a year  
- Recently visited Kagoshima with her husband  
- Buys local specialties when traveling  
- Would like to visit Shikoku  
- Uses a travel agency
\end{verbatim}
\end{footnotesize}
\end{mdframed}

\caption{Additional examples of generated personas (1).}
\label{addtional_personas1}
\end{figure}

\begin{figure}[t]
APM High:
\begin{mdframed}[linewidth=1pt]
\begin{footnotesize}
\begin{verbatim}
- Name: Miho Uchida
- 38 years old
- Female
- Lives in Hyogo Prefecture
- Lives with her husband and their daughter, 
  who is in the first grade of elementary school
- Nurse
- Cheerful and a good listener
- Goes on a trip once a year
- Recently visited Kyoto
- Enjoys taking photos while traveling
- Would like to visit Sapporo
- Makes reservations online
\end{verbatim}
\end{footnotesize}
\end{mdframed}

APM Low:
\begin{mdframed}[linewidth=1pt]
\begin{footnotesize}
\begin{verbatim}
- Name: Kana Kawamura
- 26 years old
- Female
- Lives in Tokyo
- Single and lives alone
- Freelance illustrator
- Has a free-spirited and creative personality
- Travels frequently
- Recently visited Kyoto alone
- Enjoys visiting art museums while traveling
- Would like to visit Florence
- Makes reservations based on information 
  found on social media
\end{verbatim}
\end{footnotesize}
\end{mdframed}

\caption{Additional examples of generated personas (2).}
\label{addtional_personas2}
\end{figure}

\end{document}